\pdfoutput=1

\documentclass[11pt]{article}

\usepackage{EMNLP2022}

\usepackage{times}
\usepackage{latexsym}

\usepackage{graphicx}
\usepackage{multirow}
\usepackage{amssymb}
\usepackage{amsmath}
\usepackage{multicol}
\usepackage{caption}
\usepackage{subcaption}
\usepackage{algorithm}
\usepackage{CJKutf8}
\usepackage{algpseudocode}
\usepackage{listings}
\usepackage{float}
\usepackage{times}
\usepackage{booktabs}
\usepackage{fancyvrb}
\usepackage{fdsymbol}
\usepackage{float}
\usepackage{xcolor}
\usepackage{pifont}
\newcommand{\cmark}{\ding{51}}%

\usepackage[T1]{fontenc}

\usepackage[utf8]{inputenc}

\usepackage{microtype}

\usepackage{inconsolata}

\usepackage[colorinlistoftodos,prependcaption,textsize=tiny]{todonotes}
\colorlet{punct}{red!60!black}
\definecolor{background}{HTML}{EEEEEE}
\definecolor{delim}{RGB}{20,105,176}
\colorlet{numb}{magenta!60!black}

\definecolor{codegreen}{rgb}{0,0.6,0}
\definecolor{codegray}{rgb}{0.5,0.5,0.5}
\definecolor{codepurple}{rgb}{0.58,0,0.82}
\definecolor{backcolour}{rgb}{0.95,0.95,0.92}
\lstdefinestyle{mystyle}{
    commentstyle=\color{codegreen},
    keywordstyle=\color{magenta},
    numberstyle=\tiny\color{codegray},
    stringstyle=\color{codepurple},
    basicstyle=\ttfamily\tiny,
    breakatwhitespace=false,         
    breaklines=true,                 
    captionpos=b,                    
    keepspaces=true,                 
    numbers=left,                    
    numbersep=5pt,                  
    showspaces=false,                
    showstringspaces=false,
    showtabs=false,                  
    tabsize=2
}

\title{\includegraphics[width=.04\linewidth]{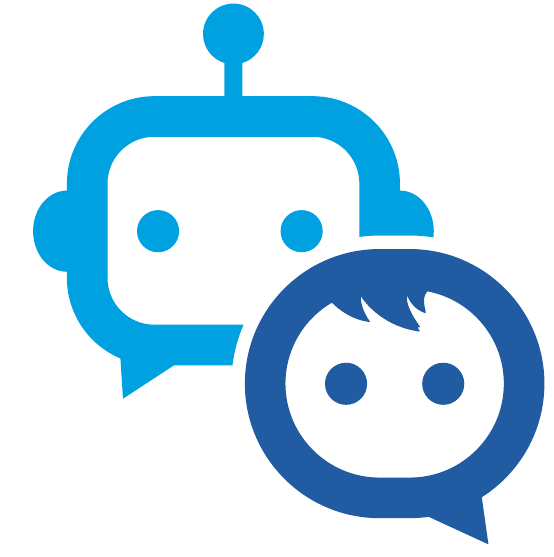}~BotSIM: An End-to-End Bot Simulation Toolkit\\ for Commercial Task-Oriented Dialog Systems}

 \author{Guangsen Wang  \quad Shafiq  Joty \quad Junnan Li  \quad Steven C.H. Hoi\\
        Salesforce Research\\
       \texttt{\{guangsen.wang, sjoty, junnan.li, shoi\}@salesforce.com}}

\begin{document}
\maketitle
\begin{abstract}
We introduce BotSIM, a modular, open-source \textbf{Bot SIM}ulation environment with dialog generation, user simulation and conversation analytics capabilities. 
BotSIM aims to serve as a one-stop solution for large-scale data-efficient end-to-end evaluation, diagnosis and remediation of commercial task-oriented  dialog (TOD) systems to significantly accelerate commercial bot development and evaluation, reduce cost and time-to-market.
BotSIM adopts a layered design comprising the infrastructure layer, the adaptor layer and the application layer.
The infrastructure layer hosts key models and components to support BotSIM's major functionalities via a  streamlined ``generation-simulation-remediation'' pipeline. The adaptor layer is used to extend BotSIM to accommodate new bot platforms. The application layer provides a suite of command line tools and a Web App to significantly lower the entry barrier  for BotSIM users such as bot admins or practitioners.
In this report, we focus on the technical designs of various system  components. 
A detailed case study using Einstein BotBuilder is also presented to show how to apply BotSIM pipeline for bot evaluation and remediation.
The detailed system descriptions can be found in our system demo paper~\citep{guangsen2022-botsim-demo}. 
The toolkit is available at: \emph{\textbf{\url{https://github.com/salesforce/BotSIM}}}.
\end{abstract}
\section{Introduction}

\begin{figure*}[!t]
  \centering
\includegraphics[width=16cm]{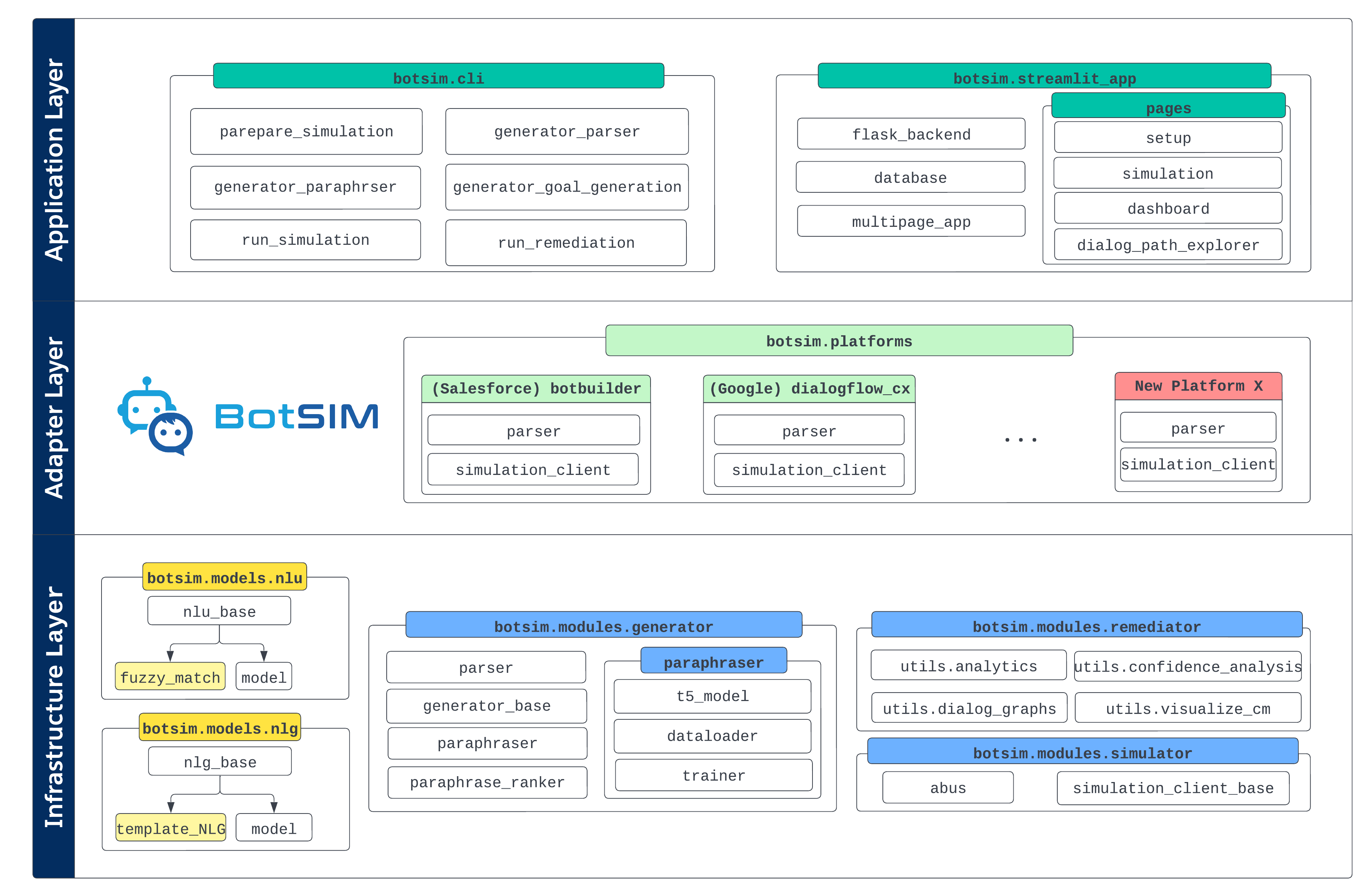}
  \caption{BotSIM System Design. }
  \label{fig:architecture}
\end{figure*}

The typical dialog system development cycle consists of dialog design, pre-deployment testing, deployment, performance monitoring,  model improvement and iteration. As in any production software system, effective and comprehensive testing at all stages is of paramount importance. 
Unfortunately, \emph{evaluating and troubleshooting} production TOD systems is still a largely manual process requiring large amount of real human conversations with the bots.
This process is time-consuming, expensive, and inevitably fails to capture the breadth of language variation present in the real world~\citep{tan-etal-2021-reliability}.  The time- and labor-intensive nature of such an approach is further exacerbated when the developer significantly changes the dialog flows, since new sets of test dialogs will need to be collected~\citep{IBM-Watson}.
Performing comprehensive end-to-end evaluation of both  NLU  and  dialog-level performance (\emph{e.g.}, task success rate) is highly challenging due to the need for further annotated efforts. Finally, there is a lack of analytical tools for interpreting  test results and troubleshooting  underlying bot issues. To address these limitations, we present \emph{BotSIM}, an open source \textbf{Bot SIM}ulation environment with dialog generation, user simulation~\citep{schatzmann-etal-2007-agenda} and conversation analytics capabilities. BotSIM aims to serve as a one-stop solution for bot practitioners  to perform  large-scale data-efficient end-to-end evaluation, diagnosis and remediation of commercial TOD systems. The modular design also allows bot developers to extend the framework to more bot platforms.

The toolkit design is depicted in Figure~\ref{fig:architecture}. It  consists of three layers: the infrastructure layer, the adaptor layer and the application layer. 
The infrastructure layer supports the upper layers through the key modules (NLU and NLG) and components (the generator, simulator and remediator). They are the ``engines'' to power  BotSIM's ``generation-simulation-remediation'' pipeline for end-to-end diagnosis, evaluation and remediation of commercial TOD bots. More details regarding the functionalities of these modules and components are given in~\citep{guangsen2022-botsim-demo}.
BotSIM currently supports two bot platforms, including Salesforce Einstein BotBuilder and Google DialogFlow CX.
The adaptor layer is designed for bot developers to extend BotSIM to new bot platforms. 
In addition to defining a set of unified  parser and client interfaces, the adaptor layer also provides some reference implementations for developers to start their own implementations of the platform-dependent parsers and chat API clients. 
The application layer offers a suite of command line tools and an easy-to-use Web App.
The Web app is designed for bot admins to directly apply BotSIM to their bots for evaluation and remediation without diving into the technical details of various system components. On the other hand, the command line tools are provided to bot practitioners for a better understanding of the toolkit to customize their own evaluation and remediation pipeline.
They can learn more about how different BotSIM components work in terms of the required inputs, expected outputs and their functionalities. 
We have provide detailed tutorials of how to use these tools in our code
\href{https://opensource.salesforce.com/botsim/latest/tutorials.html}{documentation}.

\section{Related Work}

\begin{table*}[!htbp]
\resizebox{\textwidth}{!}{%
\tiny
\begin{tabular}{ccccccccc}
\multirow{2}{*}{} &
  \multicolumn{2}{c}{Methods} &
  \multicolumn{2}{c}{Stages} &
  \multicolumn{2}{c}{Automation} &
  \multicolumn{2}{c}{Metrics} \\ \toprule 
 &
  Regression &
  End-to-end &
  Pre-deployment &
  Monitoring &
  \begin{tabular}[c]{@{}c@{}}Test case \\ curation\end{tabular} &
  \begin{tabular}[c]{@{}c@{}}User \\ Simulation\end{tabular} &
  NLU &
  \begin{tabular}[c]{@{}c@{}}Task \\ Completion\end{tabular} \\ \midrule
CX & \cmark &      &      & \cmark  &      &      &      &      \\
Watson    & \cmark  &      &      & \cmark  &      &      & \cmark  & \cmark  \\
Botium        & \cmark  &      &      &      &      &      & \cmark  &      \\
BotSIM        & \cmark  & \cmark  & \cmark  & \cmark  & \cmark  & \cmark  & \cmark  & \cmark  \\ \bottomrule
\end{tabular}%
}
 \caption{Comparison of bot evaluation capabilities of the reviewed commercial bot platforms}
 \label{tab:platform-cmp}
\end{table*}
We present the related work of BotSIM from both toolkit design and application perspectives.
\subsection{Toolkit Design}
To the best of our knowledge, there are no open source toolkits designed for simulation, evaluation and remediation of commercial chat bots. Therefore, we review some of the research toolkits/libraries designed for TOD systems and are equipped with user simulators. They include  \textbf{\texttt{ConvLab-2}}~\citep{zhu2020convlab2}, \textbf{\texttt{TC-Bot}}~\cite{LiLDLGC16-TCBot,li2017end} and  \textbf{\texttt{user-simulator}}~\cite{shi2019build}. 
\begin{itemize}
    \item \textbf{\texttt{ConvLab-2}}~\footnote{\url{https://github.com/thu-coai/ConvLab-2}} is a research toolkit for building TOD systems with state-of-the-art dialog  models. It was used in DSTC9 Track 2~\footnote{\url{https://convlab.github.io/index.html}} for the multi-domain TOD challenge. In addition, it also can perform end-to-end evaluation, and diagnose the weakness of the research systems. The essential differences to BotSIM include 1) the toolkit is limited to research TOD systems to support research models and datasets; 2) the evaluation is based on the annotated dialogs.
    \item \textbf{\texttt{TC-Bot}}~\footnote{\url{https://github.com/MiuLab/TC-Bot}} is another research-oriented framework for building end-to-end neural-based TOD systems.\textbf{\texttt{TC-Bot}} utilises a dialog-act level agenda-based user simulator to interact with the neural-based end-to-end dialog agent for policy training. Some of our simulator rules and designs are inspired and adapted from a simpler version of the \textbf{\texttt{TC-Bot}} called  \textbf{\texttt{GO-Bot-DRL}}~\footnote{\url{https://github.com/maxbren/GO-Bot-DRL}}.
    \item The ``\textbf{\texttt{user-simulator}}''~\footnote{\url{https://github.com/wyshi/user-simulator}}  provides  implementations and comparisons of different user simulators, such as the agenda-based user simulator and neural-based user simulator based on Sequicity~\citep{lei-etal-2018-sequicity}. The work aims to provide an evaluation framework for user simulator study in terms of their impacts on the trained dialog policies.
\end{itemize}
 The focus of these research toolkits is to build an end-to-end (neural-based) dialog agent prototype constrained to certain domains (\emph{e.g.}, movie booking, restaurant domain of the MultiWOZ~\citep{multiwoz} dataset). 
 On the contrary, BotSIM is designed as a holistic bot simulation environment
 for task-agnostic end-to-end evaluation, diagnosis and remediation of commercial bots. The agenda-based user simulator is only a small part of the toolkit as shown in Figure~\ref{fig:architecture}. It is designed for bot practitioners to significantly accelerate commercial bot development and evaluation, reducing human efforts, cost and time-to-market. 
 
\subsection{Commercial Bot Evaluation}
The overall comparison of different platforms in terms of testing capabilities is given in Table~\ref{tab:platform-cmp}. The detailed system reviews can be found in the system demo paper~\citep{guangsen2022-botsim-demo}.
Most current commercial platforms only focus on the regression testing rather than NLU or task-completion metrics.
While regression testing is important to ensure correct and consistent system behaviours,  it is also vital to perform pre-deployment evaluation to avoid poor user adoption and retention rate. Although some of the platforms offer turn-level NLU performance metrics, they require human efforts in curating or annotating a large number of test cases. In addition, the NLU metrics do not directly translate to goal completion performance. 
On the other hand, BotSIM can help circumvent these limitations via large scale automatic dialog generation and simulation.
\section{BotSIM Design} 
In this section, we present more details of the BotSIM design in Figure~\ref{fig:architecture}. 
The key design principles of BotSIM include modularity, extensibility and usability. These principles allow BotSIM to be adopted both by developers as a framework and bot end-users as  an easy-to-use application. 
To achieve these, BotSIM adopts a layered design comprising the infrastructure layer, the adaptor layer and the application layer. 
\subsection{Infrastructure Layer}
As the name suggests, the infrastructure layer is designed to offer fundamental model support for the framework. BotSIM's ``generation-simulation-remediation'' pipeline is powered by the models and components that reside in this layer. The models include
the natural language understanding (NLU), natural language generation (NLG) models and the key components include the generator, the simulator and the remediator.

    \textbf{\texttt{botsim.models} } package hosts BotSIM's  NLU and NLG models. From a dialogue system perspective, BotSIM can be viewed as a counterpart to a chatbot as shown in Figure~\ref{fig:simulation}: it needs to ``understand'' chatbot messages (NLU) and ``respond'' in natural languages (NLG) to carry on the conversation. Currently, fuzzy matching-based NLU and template-based NLG models are provided for efficiency reasons. More advanced  NLU and NLG models can also be incorporated by the developers by following the recipes given in the code \href{https://opensource.salesforce.com/botsim//latest/index.html}{ documentation}.  

    \textbf{\texttt{botsim.modules} } consists of the three key components to power BotSIM's ``generation-simulation-remediation'' pipeline. 
    \begin{itemize}
        \item  \texttt{botsim.modules.generator} provides two major functionalities: 1) the \texttt{parser} parses the input bot designs in the form of either metadata (Einstein BotBuilder) or API (DialogFlow CX) to infer the dialog-act maps (BotSIM's NLU); 2) the large pre-trained language model based \texttt{paraphraser} generates paraphrases from the input intent utterances. These paraphrases are used as intent queries in the simulation goals to probe bots' intent models, which allows BotSIM to perform large-scale data-efficient bot evaluation even before bots are deployed.
        \item  \texttt{botsim.modules.simulator} implements the dialog-act level agenda-based user simulation in \texttt{abus}. It also defines a simulation API client interface \texttt{simulation\_client\_base}.
        \item \texttt{botsim.modules.remediator} analyzes the simulated dialogs and produces the performance metrics and conversational analytics to support the dashboard visualisation. These metrics include both the end-to-end performance such as the task completion rates and the NLU performance. It also  offers a suite of analytical tools to provide actionable insights to troubleshoot and improve the current systems. Such tools include confusion matrix analysis and visualisation (\texttt{analytics, visualization\_cm}), dialog design graph exploration (\texttt{dialog\_graphs}).
    \end{itemize}

\subsection{Adaptor Layer}
The adaptor layer is designed for bot developers to extend BotSIM to new bot platforms. To cover new bot platforms,  the following two most important platform-specific modules of the layer must be implemented. We also provided more details in the code \href{https://opensource.salesforce.com/botsim//latest/advanced_usage.html}{documentation}. To accommodate a new bot platform, create a new package under \texttt{botsim.platforms} and implement the following classes.

    \textbf {\texttt{parser}}  acts as an ``adaptor'' to unify bot definitions (\emph{e.g.} conversation flows, intents/tasks) from different platforms to a common representation of dialog act maps. The dialog act maps are used as BotSIM NLU to map  bot messages to dialog acts. The rational is that most commercial TOD bots follow a ``rule-action-message'' design scheme and there exist clear mappings from system messages to rules/actions. For example, according to bot definitions, ``May I get your email?'' (bot message) is used to ``Collect'' (action) the ``Email'' (slot) with  entity type  ``Email'' from the user. One of the most important parser function is to parse input bot definitions (\emph{e.g.}, MetaData, API) to capture such mappings, infer the ``request\_Email'' dialog act and add the bot message to the mapping candidates of the dialog act. An example of the dialog act map is given in Figure~\ref{fig:dialog_act_map}. The major parser functions are listed in the code snippet below:
\begin{lstlisting}[language=Python]
class Parser:
  """ Parser interface"""
  def __init__(self):
    # The aggregated dialog act maps to be used as 
    # BotSIM NLU
    self.dialog_act_maps = {}
    # mapping from dialogs to their entities
    self.dialog_ontology = {}
    # conversation graph data to support path explorer
    self.conv_graph_visualisation_data = {}
  def extract_local_dialog_act_map(self):
  def conversation_graph_modelling(self, local_dialog_act_maps):
    """Model local dialog act maps and their transitions as a graph.
    :param local_dialog_act_maps: local dialog act maps (one per dialog) obtained from the parser (extract_local_dialog_act_map)
    :return:
      dialog_act_maps: aggregated dialog act maps obtained via graph traversal. These   maps need to be reviewed/revised by users before used as BotSIM NLU
      conv_graph_visualisation_data: conversation graph  data to support dialog path visualisation
        """
  def extract_ontology(self):
\end{lstlisting}
    Note the implementations of the parsers are highly platform-dependent. They require
    the developers to have access to bot platform, design and API documents to understand how bots are designed and how user inputs are elicited by the bots. There is also need to constantly revisit the implementations to incorporate new features that might be missed in the current implementations.
  We have provided our current implementations of BotBuilder and DialogFlow CX parsers for references under \texttt{botsim.platforms.botbuilder} and \texttt{botsim.platforms.dialogflow\_cx}.
    
\begin{figure*}[!t]
  \centering
  \includegraphics[width=16cm]{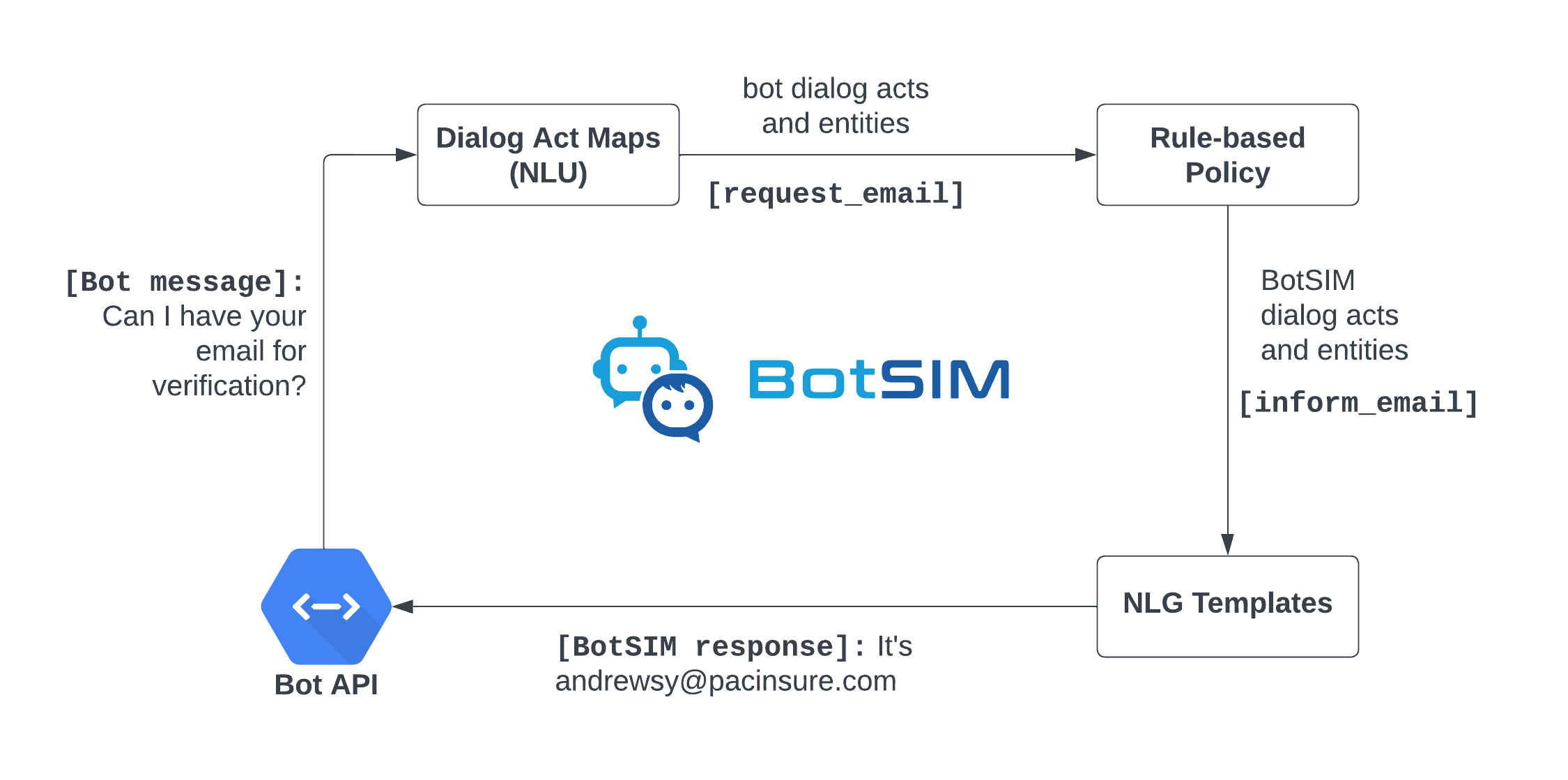}
  \caption{An exchange of dialog turns between bot and BotSIM during dialog simulation}
  \label{fig:simulation}
\end{figure*}

    \textbf {\texttt{simulation\_client}} is the other platform-dependent component for BotSIM to exchange conversations with bots via API calls. 
\begin{lstlisting}[language=Python]
from botsim.modules.simulator.user_simulator import UserSimulator
class UserSimulatorClientInterface:
    def __int__(self, config):
        self.config = config
        self._prepare_simulation()
    def _prepare_simulation(self):
    """prepare simulation according to the simulation configuration"""
    def perform_batch_simulation(self, simulation_goals, dialog_name, start_episode, config, batch_size=25):
    """perform a batch of dialog user simulations using simulation_goals
       :param simulation_goals: list of all simulation goals
       :param dialog_name: name of the dialog under simulation
       :param start_episode: starting goal index of simulation_goals
       :param config: simulation settings including information such as API tokens
       :param batch_size: number of simulation sessions per batch"""
        user_simulator = UserSimulator(simulation_goals, config)
        ...
\end{lstlisting}
    Figure~\ref{fig:simulation} depicts how a dialogue turn between the bot and BotSIM is conducted via bot APIs.
    Using dialog act maps as NLU, a rule-based dialog state manager (policy implemented in \texttt{botsim.modules.simulator}) takes in the bot dialog acts and produces the corresponding user dialog acts. The user dialog acts are converted to natural language responses by the template-based NLG. The natural language responses are sent back via API. The conversation ends when the task has been successfully finished or an error has been captured.
    Similar to the parser, the implementation of the client is also highly platform-dependent. Developers can refer to our implementations for BotBuilder and DialogFlow CX when extending BotSIM to new bot platforms.

\subsection{Application Layer}
The application layer is designed to significantly flatten the learning curves of BotSIM for both bot developers/practitioners and end-users. 

    \textbf {Command Line Tools} \texttt{botsim.cli} contains a set of command line tools for practitioners to learn more about the  major BotSIM components. The ``generation-simulation-evaluation'' pipeline has been split into multiple stages to expose the required inputs and expected outputs for each stage. 
    They serve as basic building blocks for bot practitioners to build their customized pipelines or apply only certain tasks rather than the whole BotSIM pipeline.
\begin{lstlisting}[language=Bash]
python botsim/cli/run_generator_parser.py \
       --platform $platform \
       --test_name $name
python botsim/cli/run_generator_paraphraser.py \
       --platform $platform \
       --test_name $name
python botsim/cli/run_generator_goal_generation.py \
       --platform $platform \
       --test_name $name
python botsim/cli/run_simulator.py \
       --platform $platform \
       --test_name $name
python botsim/cli/run_remediator.py \
      --platform $platform \
      --test_name $name
\end{lstlisting}
    
    \textbf {Streamlit Web App} \texttt{botsim.streamlit\_app} is a multi-page easy-to-use Web app for end users such as bot admins without diving into  technical details.
    The app can be built as a docker image  and deployed to various cloud platforms (\emph{e.g.} GCP) for access to more computation resources. We use Streamlit~\footnote{\url{https://streamlit.io/}} to build the front-end pages. Flask is used to implement the backend APIs for Streamlit to invoke BotSIM functionalities. The app is also equipped with a SQL database to keep track of  simulation stages and simulation performance. BotSIM supports two types of SQL databases including Sqlite3 and Postgres. 

\section{Einstein BotBuilder Case Study}

In this section, we show how to perform end-to-end evaluation and remediation of the pre-built ``Template Bot'' from the Salesforce Einstein BotBuilder platform. We follow BotSIM's ``generation-simulation-remediation'' pipeline as detailed in~\citep{guangsen2022-botsim-demo}.
The ``Template Bot'' has six dialog intents. Each intent has a set of  hand-crafted training utterances. 
For controllable experiments, we  sample 150 utterances per dialog as the training set (train-original) and use the remaining for evaluation (eval-original).  The six intents are: ``Transfer to agent (TA)'', ``End chat (EC)'', ``Connect with sales (CS)'', ``Check issue status (CI)'', ``Check order status (CO)'' and ``Report an issue (RI)''.  The dataset information is given in Table~\ref{templatebot_dataset}.
\begin{table*}[!t]
\centering
\begin{tabular}{cccccccc}
Dataset & Intent enquiries & TA  & EC  & CS  & CI  & CO  & RI   \\ 
\toprule
\multirow{2}{*}{Train} 
& train-original & 150 & 150 & 150 & 150 & 150 & 150  \\
& train-augmented  & 255  &  184 &  212  &  268   & 215  &  294 \\ 
\midrule
Dev & train-paraphrases & 1465   &  1467 &  1754  &  1989  & 1895  &  1786   \\ 
\midrule
\multirow{2}{*}{Eval}  & eval-original   &   182 & 145   &  183  &  222  & 205   & 178 \\
& eval-paraphrases   &  1190  & 933   &  1648  & 2172   &    1936 &   1795\\
\bottomrule
\end{tabular}
\caption{Dataset information for the Einstein Template Bot case study. }
\label{templatebot_dataset}
\end{table*}

\subsection{Parse bot metadata}
The required inputs for BotSIM include 1) bot design metadata containing the bot designs (\emph{e.g.}, intents/dialogs, entities) 2) intent utterance metadata 3) LiveAgent API information. They can be retrieved from users’ Salesforce org.  
BotSIM starts by parsing the input metadata and generates the NLU (Dialog Act Maps) and NLG (Response) templates needed for dialog simulation. 
\subsection{Revise dialog act maps and ontology}
\begin{figure*}[t]
  \centering
  \includegraphics[width=16cm]{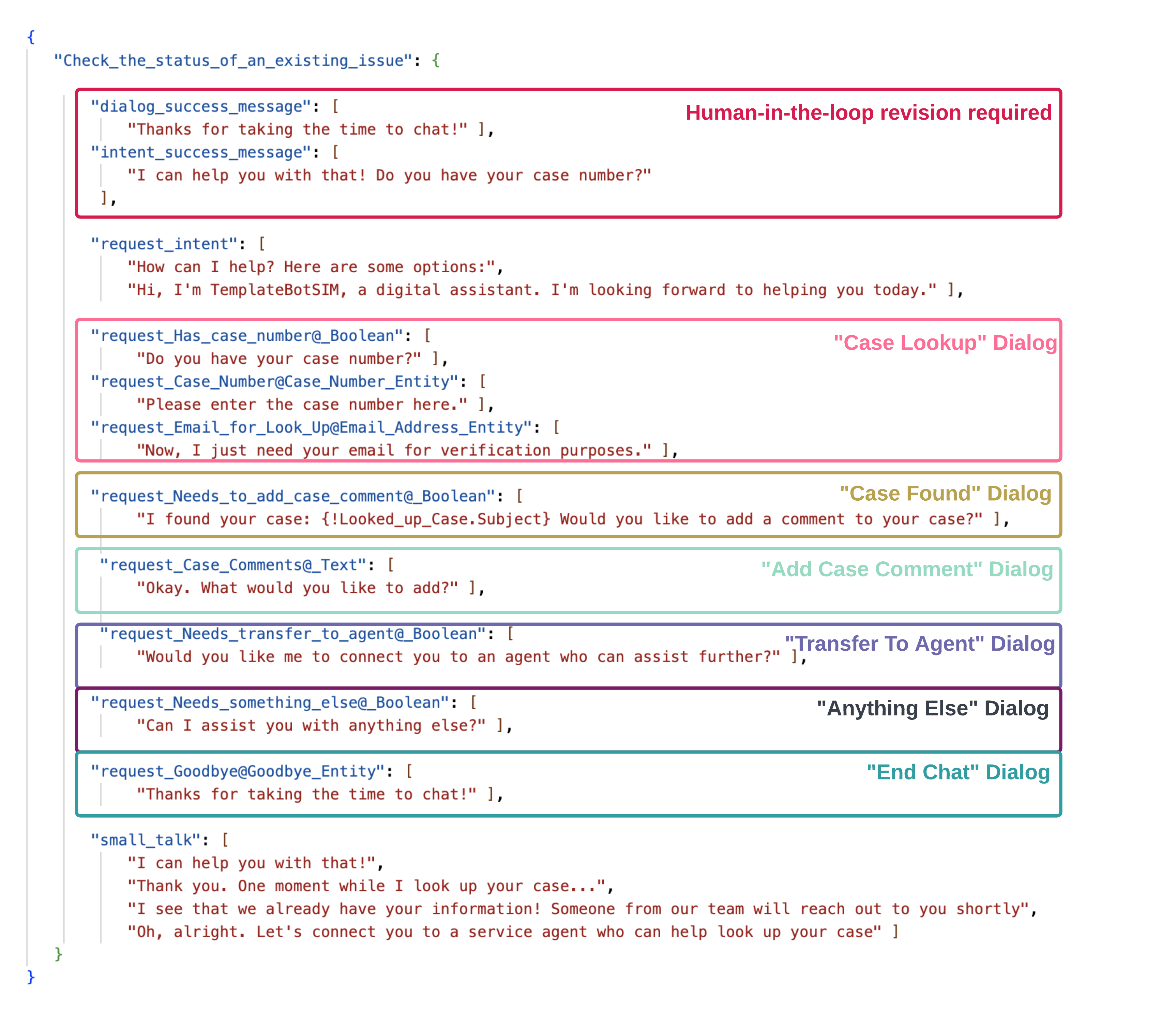} 
  \caption{Revised dialog act map for dialog intent ``Check the status of an existing issue'' of Einstein BotBuilder}
  \label{fig:dialog_act_map}
\end{figure*}

The generated NLU dialog act maps convert bot messages to dialog acts via fuzzy matching during dialog simulation. An example of dialog act maps is shown in Figure~\ref{fig:dialog_act_map}. 
The aggregated dialog act map is inferred automatically by the parser by modelling the bot design as graphs: 1) individual dialogs are firstly parsed to get their ``local'' dialog act maps; 2) individual dialogs (including sub-dialogs such as ``Case Lookup'')  are modelled as vertices  associated with  their ``local'' dialog act maps; 3) dialog transitions of the entire bot are modelled as graph edges. The final aggregated dialog act map of a dialog is created by collecting all the ``local'' dialog act maps along the paths  starting from the dialog node to the successful dialogs (\emph{e.g.,} ``End\_Chat''). Meanwhile, the parser also extracts the bot entity ontology. It lists all entities and their randomly initialized values for each dialog.

As the only ``human-in-the-loop'' step in the BotSIM pipeline, review or revision of the automatically inferred dialog act maps and ontology are required by BotSIM users. For dialog act maps, the two special dialog acts,  ``dialog\_success\_message'' and ``intent\_success\_message'',   are the  golden labels  indicating a successful task completion and a correct intent classification, respectively. They are inferred heuristically by regarding the first dialog message  as  the ``intent\_success\_message'' and the last message as the ``dialog\_success\_mesage''. 
Users are required to verify these two dialog acts for each evaluation dialog to ensure their correctness. Note the review of dialog act maps is a one-time effort unless there are significant changes made to the bot design.
Since the entity values of the ontology are mostly related to users' products and services, the randomly initialised values in the ontology file may be replaced with real ones in order to evaluate the named entity recognition (NER) model. 
The revised dialog act maps and the ontology are subsequently used to create the simulation goals as in Figure~\ref{goal} for dialog simulation.

\subsection{Paraphrase and goal generation}
Agenda-based user simulation requires pre-defined goals to ensure that the simulated user behaves in a consistent, goal-directed manner~\cite{schatzmann-etal-2007-agenda}. The goals consist of a set of constraints ( dialog acts and entity slot-value pairs) needed to complete the task. An exemplar simulation goal for ``check the status of an existing issue'' is given below:
\begin{lstlisting}[label={goal},language=Python]
{
   "Goal": {
     "Check_the_status_of_an_existing_issue_0": {
       "inform_slots": {
         "Has_case_number@_Boolean": "no",
         "Case_Number@Case_Number_Entity": "P3V4S6",
         "Email_for_Look_Up@Email_Address_Entity": "andrewadams@example.org",
         "intent": "I would like to know about my issue.",
         "Needs_transfer_to_agent@_Boolean": "no",
         "Needs_something_else@_Boolean": "no",
         "Goodbye@Goodbye_Entity": "goodbye",
         "Needs_to_add_case_comment@_Boolean": "no",
         "Case_Comments@_Text": "P94RMU"
       },
       "request_slots": {
         "Check_the_status_of_an_existing_issue": "UNK"
       },
       "name": "Check_the_status_of_an_existing_issue"
     }
   }
}
\end{lstlisting}
Note how the entity slots are taken from the dialog act maps in Figure~\ref{fig:dialog_act_map}. In particular, the ``intent'' slot contains the intent query to test the intent models. To enable pre-deployment testing  without any human-written test cases, we apply the paraphrasing models to the ``train-original'' utterances to get the ``train-paraphrases'' dataset.  
Compared to the number of original utterances, the ``train-paraphrases'' dataset is roughly ten times larger as shown in Table~\ref{templatebot_dataset}.
Simulation goals are created by taking the ``train-paraphrases'' as the intent queries to simulate the variations in real user intent queries. The ``train-paraphrases'' goals are subsequently  used as the development set to perform end-to-end evaluation of the dialog system via dialog simulation. 
\subsection{Dialog simulation results}
In this study, we focus on the intent model for two reasons. Firstly, the intent model  can be retrained. Secondly, wince we do not have any customer data,  the entity values in the goals are randomly generated and may not reflect the real-world values.  
After dialog simulation with ``train-paraphrases'' goals, the original intent training sets are augmented with the wrongly classified intent queries (``train-augmented'') to re-train the intent model according to the remediation suggestions. 
We then compare the performance before and after retraining on the goals created from the human-written ``eval-original'' utterances in Table~\ref{f1-comparison} in terms of intent F1 scores.

\begin{table*}[!t]
\centering
\begin{tabular}{cccccccc}
Model  & Eval. & TA & EC & CS & CI & CO & RI   \\ 
\toprule
\multirow{2}{*}{Baseline}  
& original   &  0.92$\pm$0.03  &  0.95$\pm$0.02  &  0.89$\pm$0.03  & 0.93$\pm$0.03   & 0.94$\pm$0.02 &   0.82$\pm$0.04  \\ 
& paraphr.  &  0.88$\pm$0.01  & 0.93$\pm$0.01   & 0.85$\pm$0.01   & 0.91$\pm$0.01  &  0.93$\pm$0.01  & 0.77 $\pm$0.02      \\
\midrule
\multirow{2}{*}{Retrained} 
 & original   & 0.92$\pm$0.03  & 0.97$\pm$0.02   &  0.93$\pm$0.03  & 0.95$\pm$0.02  & 0.96$\pm$0.02 &  0.87$\pm$0.04     \\
 & paraphr.  & 0.89$\pm$0.01  &  0.94$\pm$0.01 & 0.90$\pm$0.01 & 0.94$\pm$0.01  & 0.94$\pm$0.01 &0.80$\pm$0.02    \\
\bottomrule
\end{tabular}
\caption{Results for the Einstein Bots case study, before and after retraining the intent model with the augmented training set (F1 with 95\% confidence interval computed with 10K bootstrapped samples).}
\label{f1-comparison}
\end{table*}

We observe consistent improvements for all intents after model retraining. The more challenging the intents (lower F1s) are, \emph{e.g.,} ``Report an issue'' and ``Connect with sales'', 
the larger performance gains are observed compared to the easier intents such as ``End Chat'' (higher  F1s). This demonstrates the efficacy of BotSIM in intent model improvement. 
We conjecture the improved performance of the more challenging intents is due to more paraphrases being selected for retraining to better cover the language variation. 
\subsection{Diagnosis and Remediation Dashboard}
The remediator generates health diagnosis reports, performs analyses, and provides actionable recommendations to troubleshoot and improve dialog systems.
The major dashboard components are presented in Figure~\ref{fig:remediator-reports}, \ref{fig:remediator-intent}, \ref{fig:remediator-ner}, \ref{fig:remediator-analytics}.
 \paragraph{Bot health reports.} The bot ``health'' dashboard consists of a set of multi-scale  performance reports. At the highest level, users can have a historical view of most recent simulation/test sessions (\emph{e.g.,} after each major bot update). The historical performance comparison can help users  evaluate the impacts of bot changes quantitatively, from which they can make decisions like whether or not keep certain changes. 
 In the test session performance summary report, users can have the details of a selected test session including the data distribution, overall dialog performance metrics across all dialog intents of the test session. The dialog-specific performance report provides detailed  detailed intent and NER performance of the selected dialig intent. Through the dialog-specific performance report, one can quickly identify the most confusing intents and entities. This saves significant efforts and helps better allocation of resources for troubleshooting and bot improvement.
 \paragraph{Remediation recommendations.} 
 In addition to the diagnosis reports, the remediator also provides actionable insights/suggestions for users to remedy some of the identified issues. 
 The root causes of the failed conversations are identified via backtracking of the simulation agenda.
The recommendation dashboards (Figure~\ref{fig:remediator-intent} and Figure~\ref{fig:remediator-ner}) allow detailed investigation of all intent or NER errors along with their corresponding simulated chat logs. 
For intent models, the paraphrase intent queries that lead to intent errors are grouped by the original intent utterances. These original intent utterances are sorted by the number of intent errors of their paraphrases in descending order (drop-down list of the Remediation suggestions in  Figure~\ref{fig:remediator-intent}). Depending on the wrongly classified intents, the remediator would suggest some follow-up actions. For example, 1) augmenting the intent training set with the queries deemed to be out-of-domain by the current intent model, 2) moving the intent utterance to another intent if most of  paraphrases of the former intent  utterance are classified to the latter intent. Note the suggestions are  meant to be used as guidelines rather than strictly followed. More importantly, they can always be extended by users to include domain expertise in troubleshooting bots related to their products/services.
 \paragraph{Conversation analytics.} Another useful component of the Remediator is the suite of conversation analytical tools as shown in Figure~\ref{fig:remediator-analytics}. They further help bot practitioners gain more insights for troubleshooting and improving their dialog systems. The confusion matrix analysis breaks down the intent model performance into (sortable) recall, precision and F1 accuracies to help identify the worse performing intents. It also detects potential intent overlaps  using the clustering algorithms in ~\citep{thoma2017analysis} based on the performance metrics. Another useful analytical tool is the tSNE~\citep{JMLR:v9:vandermaaten08a} clustering of the intent utterances using sentence transformer~\citep{reimers2019sentencebert}  embeddings. The  tSNE visualisation enables users to  gauge the training data quality. It is also an effective tool in identifying overlapping intents and can potential benefit new intent discovery as well. 
 \paragraph{Dialog path explorer} Lastly, powered by parsers' conversation graph modelling capability, the dialog path explorer can be used to visualise different dialog paths of the current bot design. For example, users can select the ``source'' and ``target'' dialogs and explore the generated dialog paths. Not only is the tool valuable for comprehensive testing coverage of conversation paths, it also offers a controllable approach to troubleshooting dialog design related errors or even helping improve the current bot design.

\section{Conclusion}
We presented the design of BotSIM,  an open source, modular end-to-end bot simulation toolkit for evaluation, diagnosis and  remediation of commercial TOD systems. Through the streamlined ``generation-simulation-remediation`` pipeline, BotSIM can be adopted to accelerate commercial bot development and evaluation, reduce cost and time-to-market.
Thanks to the layered design, not only can it be used as a framework for bot developers to extend BotSIM to new bot platforms, it also offers a suite of easy-to-use command line tools and Web App for bot practitioners to directly apply BotSIM to their bots.
We have open-sourced the \href{https://github.com/salesforce/BotSIM}{toolkit} at \emph{\textbf{\url{https://github.com/salesforce/BotSIM}}} including the Streamlit Web App. We also provides detailed \href{https://opensource.salesforce.com/botsim//latest/index.html}{documentations} to accompany the code.
We welcome contributions from the community to help improve and extend BotSIM further.

\section{Limitations}
For efficiency reasons, BotSIM adopts a template-based NLG model for converting user dialog acts to natural languages. Although the template-NLG is more controllable and flexible compared to the model-based NLG, they may lack naturalness. One possible future improvement includes a combination of template-based NLG and the model-based NLG. For example, we can train a model-based NLG to generate templates~\citep{DBLP:journals/corr/abs-1808-10122} for BotSIM's response templates. In this way, both efficiency and naturalness can be achieved. Currently, BotSIM is trained and evaluated utilizing English text. We leave multi-lingual bot simulation capability as one of our future works.
\section{Broader Impact}
The pretrained language{-}model based paraphrasers (T5-base and Pegasus) used in this study are pretrained and finetuned with large scale of text corpora scraped from the web, which may contain biases. These biases may even be propagated to the generated paraphrases, causing harm to the subject of these stereotypes. Although the paraphrasing models are only applied to generate the testing intent queries, BotSIM users are advised to take into consideration these ethical issues and may wish to manually inspect or otherwise filter the generated paraphrases. 
It is also noteworthy that to prevent any data privacy leakage, the information produced in the simulation (the entity values in the BotSIM ontology) is randomly generated, and therefore fake. This includes the email addresses, names. 

\appendix

\begin{figure*}[t]
  \centering
\includegraphics[width=17cm]{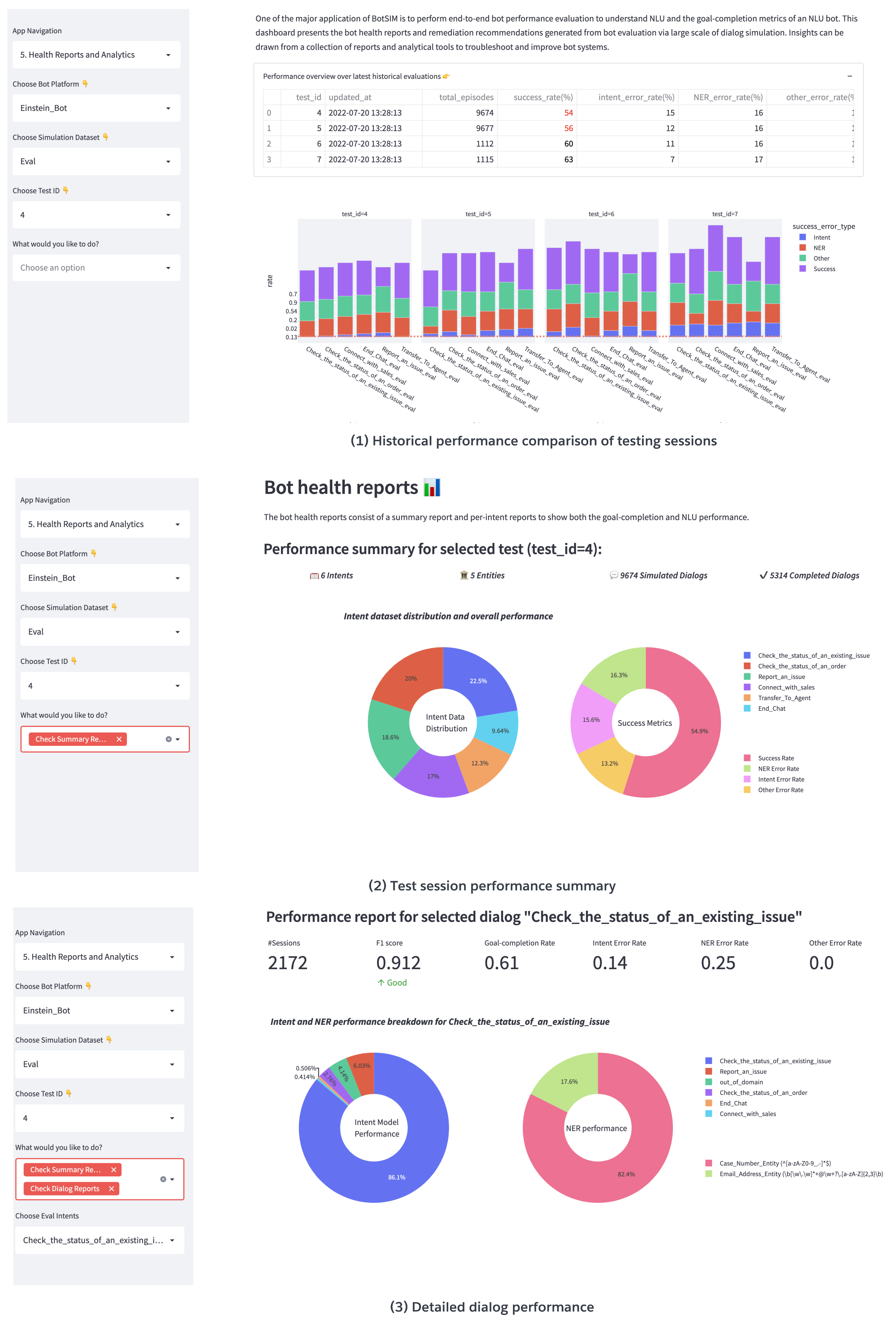} 
  \caption{Remediator dashboard: bot health reports. }
  \label{fig:remediator-reports}
\end{figure*}

\begin{figure*}[t]
  \centering
\includegraphics[width=17cm]{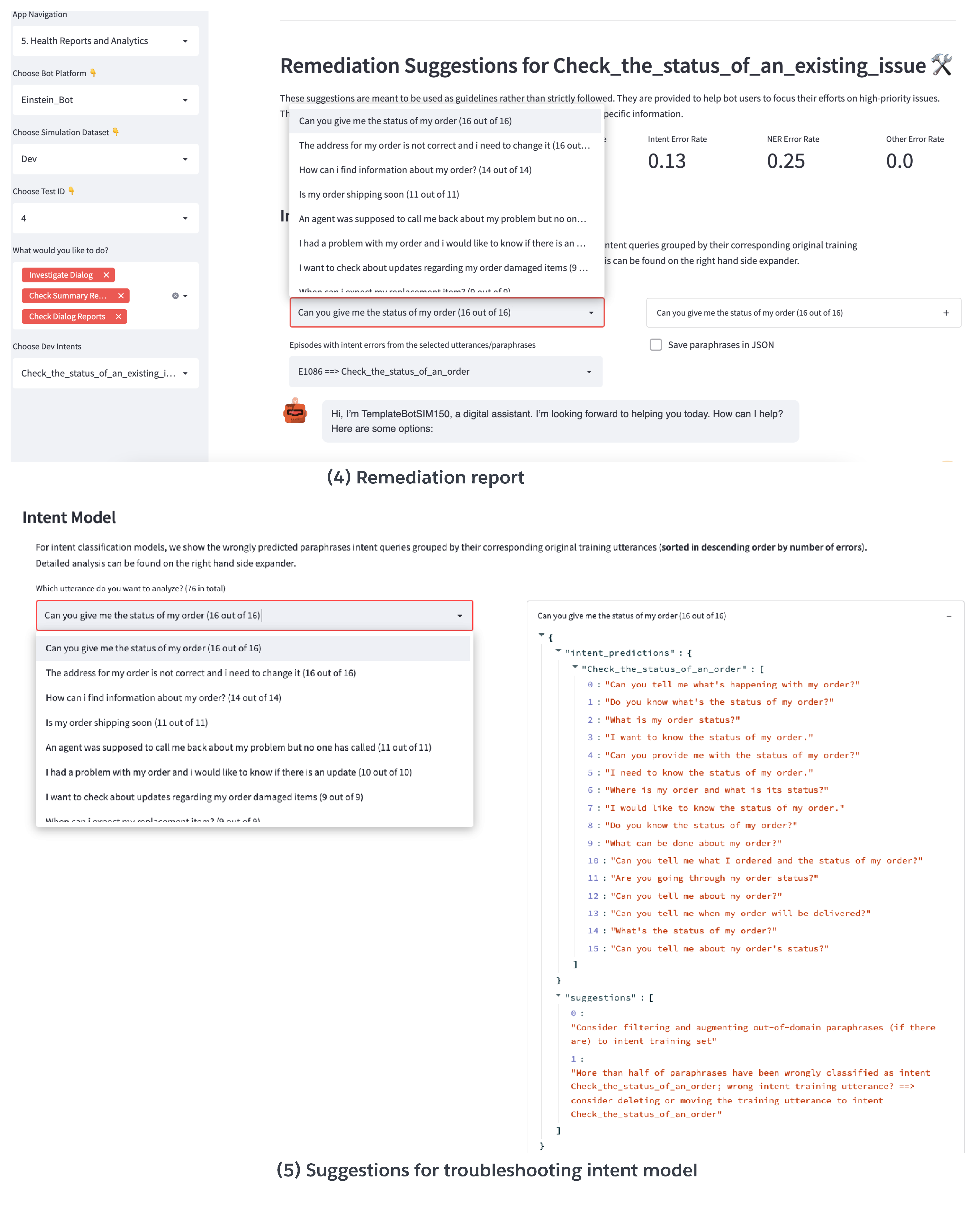} 
  \caption{Remediator dashboard: intent model remediation suggestions. }
  \label{fig:remediator-intent}
\end{figure*}

\begin{figure*}[t]
  \centering
\includegraphics[width=15cm]{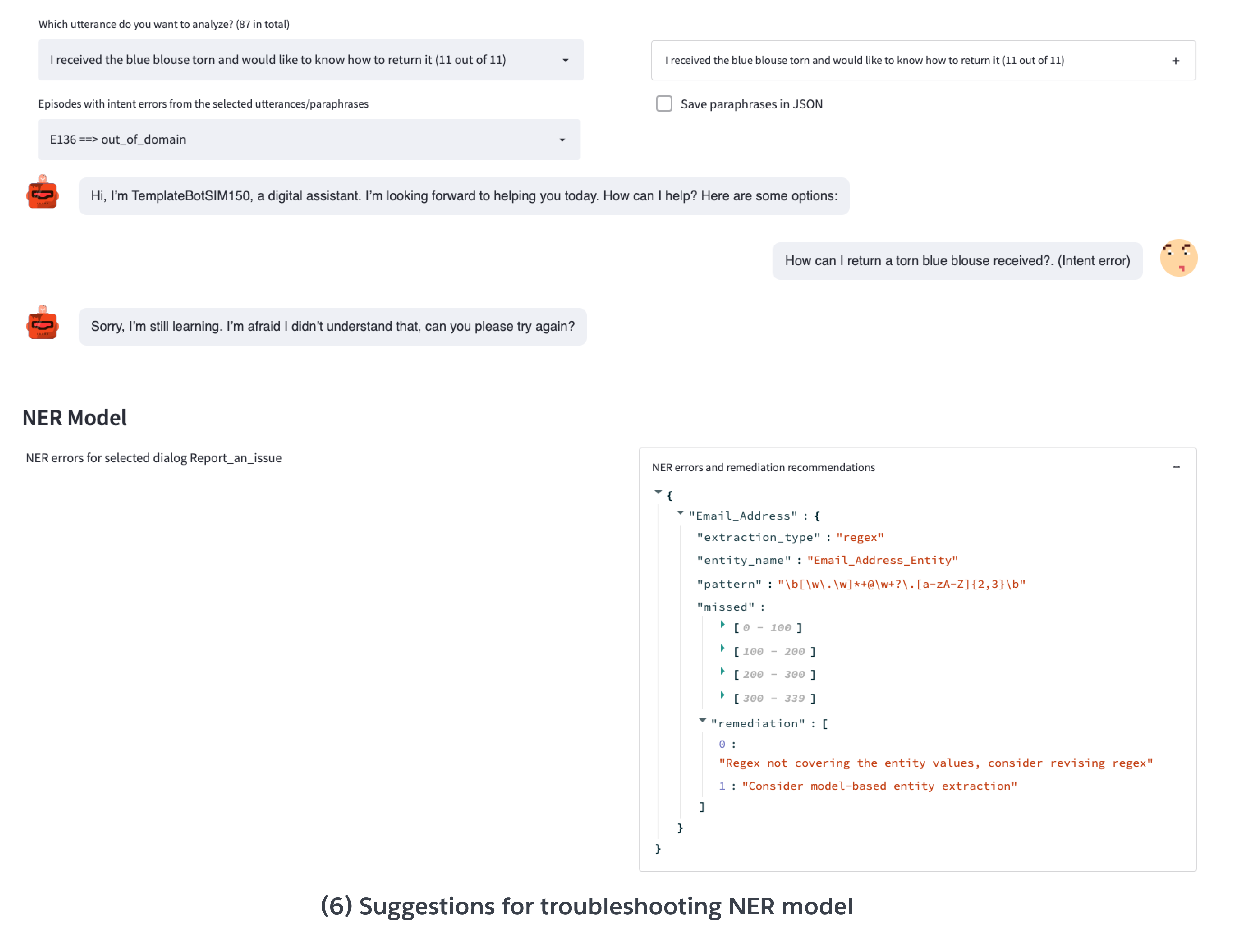} 
  \caption{Remediator dashboard: NER remediation suggestions. }
  \label{fig:remediator-ner}
\end{figure*}

\begin{figure*}[t]
  \centering
\includegraphics[width=17cm]{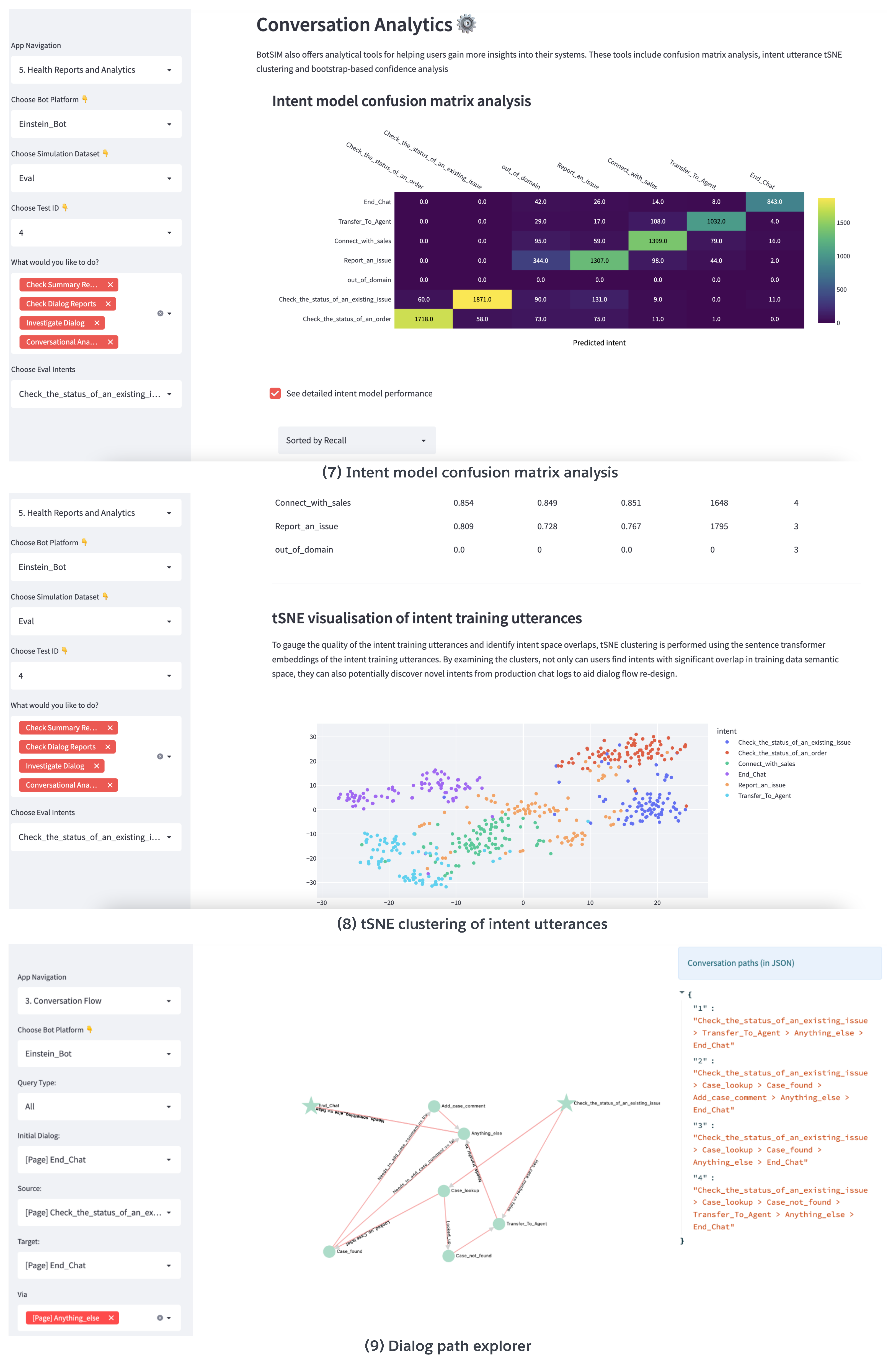} 
  \caption{Remediator dashboard: Conversational analytical tools. }
  \label{fig:remediator-analytics}
\end{figure*}

\bibliography{anthology,custom}
\bibliographystyle{acl_natbib}
\end{document}